\documentclass{article}
\usepackage{spconf,amsmath,graphicx}
\usepackage{amssymb}
\usepackage{multirow}
\usepackage{hhline}
\usepackage{xcolor}
\usepackage{subfigure}
\usepackage{booktabs}
\usepackage{hyperref}

\usepackage{pifont}

\usepackage[font=small]{caption}
\title{AL2: Progressive Activation Loss for Learning General\\ Representations in Classification Neural Networks}

\name{Majed El Helou \quad Frederike D{\"u}mbgen \quad Sabine S{\"u}sstrunk}
\address{School of Computer and Communication Sciences, EPFL, Switzerland.}

\begin{document}
\small
\maketitle

\begin{abstract}
The large capacity of neural networks enables them to learn complex functions. To avoid overfitting, networks however require a lot of training data that can be expensive and time-consuming to collect. A common practical approach to attenuate overfitting is the use of network regularization techniques. 

We propose a novel regularization method that progressively penalizes the magnitude of activations during training. The combined activation signals produced by all neurons in a given layer form the representation of the input image in that feature space. We propose to regularize this representation in the last feature layer before classification layers. Our method's effect on generalization is analyzed with label randomization tests and cumulative ablations. Experimental results show the advantages of our approach in comparison with commonly-used regularizers on standard benchmark datasets.
\end{abstract}


\section{Introduction} \label{sec:introduction}
Deep neural networks continue to achieve increasingly-better results on a wide range of tasks; medical image analysis~\cite{xu2014deep}, semantic segmentation~\cite{zhu2019improving}, finding robust features for audiovisual emotion recognition and object recognition~\cite{kim2013deep}. Improvements in the underlying hardware and in parallelization strategies~\cite{deng2012scalable,doersch2017multi} pave the way for ever larger networks. The size of such networks contributes to the complexity of the functions they can model and thus allows the learning of richer representations. However, this increase in complexity can also come at a cost. The capacity of deeper networks increases and so does their potential to memorize~\cite{arpit2017closer}. This problem drives the need for larger and more varied training datasets. Such datasets increase the training time, and are also expensive and time-consuming to collect.

To reduce overfitting and improve generalization, various regularization methods are currently used in the training of neural networks. Regularizers such as batch normalization~\cite{ioffe2015batch}, dropout~\cite{srivastava2014dropout}, and weight decay~\cite{krogh1992simple} are commonly used but are not sufficient~\cite{blot2018shade}, and the neural networks are still capable of simply memorizing an entire training set~\cite{zhang2017understanding}. Neural network regularization remains an open problem~\cite{blot2018shade,zhang2017understanding,morcos2018importance,li2019understanding}. A recent method addresses this problem by proposing to minimize the intra-class entropy of the network representations~\cite{blot2018shade}. However, the main assumption of intra-class similarity fails in the presence of incorrect labels, again requiring the costly perfectly-annotated training datasets.

We propose a simple regularization method that is applied on the feature representation learned by the neural network. This is the representation used by the final linear layers for classification. Inspired by recent findings that neural networks learn general patterns first~\cite{arpit2017closer,han2018co,ulyanov2018deep}, we propose an $\ell_2$-based activation-regularization loss (AL2) that increases per epoch to progressively regularize the network and not allow it to memorize the dataset-specific patterns that lead to overfitting. AL2 directly acts on feature representations through a loss imposed on the magnitude of their activations.

Label randomization results show that our AL2 regularization significantly improves the generalization of the baseline convolutional neural network (CNN). AL2 has a significant effect on the fundamental representation learning, as shown by our canonical correlation analysis~\cite{raghu2017svcca,morcos2018insights} in Section~\ref{subsec:CCA}. Additionally, we show that our method combines well with batch normalization, dropout, and weight decay, which can thus achieve better generalization when combined with AL2. Besides label randomization, the cumulative ablation study~\cite{morcos2018importance} results in Section~\ref{subsec:ablation} show that our CNN trained with AL2 has better generalization strength than the different baselines, even at epochs where both have roughly equal test accuracy.

Our contributions are summarized as follows. \textbf{1)} We present AL2, a progressive regularization method acting on the activations of the feature representation learned by neural networks before their final classification layers. \textbf{2)} We show that our approach improves the generalization of the learned representation: first empirically with label randomization experiments, then using a recent cumulative ablation strategy for assessing the generalization of learned representations. \textbf{3)} We analyze the effect of AL2 on the learned representation through a canonical correlation analysis.

\section{Related Work} \label{sec:related_work}
\textbf{Generalization.} Generalization in neural networks remains an open question~\cite{neyshabur2019towards,recht2019imagenet}. The sharpness or flatness of the minima found in weight optimization is commonly used to indicate, respectively, bad or good generalization~\cite{keskar2017large,hoffer2017train}. However, this belief is undermined by the fact that, for different flatness definitions, the value of flatness can be modified without modifying the function learned by the neural network~\cite{dinh2017sharp}. Even the performance in terms of error on the held-out validation or test sets is not always a perfect indicator of generalization~\cite{soudry2018implicit}. One approach to assess the quality of the feature representation learned by a network is to evaluate how much it actually memorizes. This can be achieved by training with a portion of randomized class labels, as the only way the network can learn to predict these random labels is by memorizing this data~\cite{zhang2017understanding,neyshabur2017exploring}. The result is a measure related to the empirical Rademacher complexity~\cite{bartlett2002rademacher}.

Recently, learning despite the presence of corrupt labels in the dataset has become popular~\cite{ma2018dimensionality,yu2019does,yi2019probabilistic}. These methods, however, explicitly aim to solve this noisy-learning problem, by modeling it or by re-labeling the dataset. 
This is not our objective in using corrupt labels. We use the randomization as an assessment tool of the effects of regularizers on memorization. Another assessment method we use consists of randomly ablating activation signals at inference time, and its results correlate well with generalization strength~\cite{morcos2018importance}.

\textbf{Regularization.} The most commonly-used regularization methods to reduce network overfitting are batch normalization~\cite{ioffe2015batch}, dropout~\cite{srivastava2014dropout}, and weight decay~\cite{krogh1992simple}. Batch normalization attempts to stabilize the output of one layer to aid the learning of the following one, dropout attempts to increase robustness by forcing random signal ablations during training, and weight decay reduces network complexity by penalizing the norm of some or all optimization weights. It is recently shown that batch normalization and dropout have disharmonious behaviors~\cite{li2019understanding}, as they have opposite effects on feature variance between training and inference. Network regularization remains an open problem~\cite{blot2018shade,zhang2017understanding,morcos2018importance}, along with the study of generalization.

\textbf{Representations.} Feature representations are not only important for transfer learning but also for application-specific feature extraction~\cite{xu2014deep,el2019mobile}. Canonical correlation, which we use in our representation analysis, has been recently shown to be a good distance metric to measure similarities of learned representations and to obtain more insights~\cite{raghu2017svcca,morcos2018insights}.

\section{Method}
We present a regularization method that can be applied on standard classification neural networks. The network architecture is first separated into a \emph{trunk} $\phi$ and a \emph{head} $\psi$. The trunk extracts features from the input image and creates a representation signal that is passed to the head. The head then uses the extracted features to perform its classification and predict a probability for each class. The representation learned by the trunk should focus on important image features that can generalize well to unseen data, and not simply extract data-specific patterns. It is this representation that we regularize using our AL2 loss.

The regularization loss is the norm of the feature layer's activation values, and is added as an auxiliary loss term to the classification loss. The overall loss for mini-batch $\mathcal{B}$ is then given at epoch $e$ by
\begin{equation} \label{eq:AL2}
    \begin{split}
        \mathcal{L}_e\big(x,y;\Theta\big) = \sum_{x\in \mathcal{B}}  {\mathcal{L}_c\big(\psi(\phi(x)), y; \Theta_c \big) } +
            \lambda_e {\mathcal{L}_r\big(\phi(x); \Theta_r \big) } ,
    \end{split}
\end{equation}
where $(x,y)$ are (image, label) pairs, $\Theta=\Theta_c \cup \Theta_r$ is the set of parameters over which the loss is optimized, and $e$ is a given epoch in the training. $\mathcal{L}_c$ is the classification loss, $\mathcal{L}_r$ is our activation regularization loss, $\phi(\cdot)$ is the function learned by the trunk to extract the feature representation, $\psi(\cdot)$ is the function learned by the head to perform the prediction, and $\lambda$ is a series of weights. In all our experiments, $\mathcal{L}_c$ is the cross-entropy loss, $\mathcal{L}_r$ is the $\ell_2$ norm and the series of weights $\lambda$ is defined as
\begin{equation}
    \lambda_e = \lambda_{e-1}*(1.1*u[5-\lambda_{e-1}] + 1.01*u[\lambda_{e-1}-5])
\end{equation}
$\forall e>0$, where $u[\cdot]$ is the Heaviside function. Results are not extremely sensitive to changes in this series of empirically-chosen weights as long as they are increasing with an exponential trend, even when a geometric series (only a single factor) is used. We thus use this series of weight values with $\lambda_0=0.01$ in all our experiments. Similar to weight decay or other regularizers, the parameter $\lambda$ can be tweaked for a given dataset or network architecture. The reason it is progressively increasing is that neural networks learn general patterns first, and then overfit to the data-specific patterns~\cite{arpit2017closer,han2018co,ulyanov2018deep}. Therefore, by leaving less and less flexibility to the network as the training advances, we limit its memorization capacity in later stages and minimally affect its learning phase in the earlier stages.

Our auxiliary regularization loss does not directly constrain a set of weights, whether in the trunk or in the head of the network. It only constrains the activations of the learned representation. This makes it more general than weight decay, which, in contrast, directly acts on a user-specified set of weights. In fact, weight decay has, in our experiments, the least effect on the trunk's final activation magnitudes when compared with the other regularizers. Our AL2 does, however, regularize the network, but while leaving the flexibility to use any or all of the trunk's layers to minimize this loss.

\begin{table*}[t]
\centering
\begin{tabular}{cc| *{6}{c}||c||}
\toprule
\multicolumn{9}{c}{\textbf{Different metrics evaluated across training epochs (\textcolor{gray}{without}/with AL2)}} \\ \cline{1-9}
Baseline & Metric & epoch=100 & epoch=200 & epoch=300 & epoch=400 & epoch=500 & epoch=600 & epoch=700\\  \cline{1-9} 
\multirow{3}{*}{Bare} & TA & \textcolor{gray}{84.20}/95.25 & \textcolor{gray}{45.30}/94.92 & \textcolor{gray}{25.25}/93.07 & \textcolor{gray}{23.83}/88.76 & \textcolor{gray}{26.07}/79.64 & \textcolor{gray}{26.45}/75.88 & \textcolor{gray}{25.84}/\textbf{68.46} \\
 & $\mathcal{L}_c$ & \textcolor{gray}{2.15}/2.22 & \textcolor{gray}{1.78}/2.19 & \textcolor{gray}{0.89}/2.15 & \textcolor{gray}{0.19}/2.11 & \textcolor{gray}{0.04}/2.08 & \textcolor{gray}{0.01}/2.07 & \textcolor{gray}{0.00}/2.08 \\
 & $\mathcal{L}_r$ & \textcolor{gray}{3.20}/0.24 & \textcolor{gray}{10.93}/0.10 & \textcolor{gray}{26.12}/0.06 & \textcolor{gray}{54.42}/0.03 & \textcolor{gray}{74.49}/0.02 & \textcolor{gray}{103.26}/0.01 & \textcolor{gray}{119.10}/0.00 \\
\cline{1-9} \multirow{3}{*}{BN~\cite{ioffe2015batch}} & TA & \textcolor{gray}{74.72}/95.47 & \textcolor{gray}{36.65}/94.48 & \textcolor{gray}{26.72}/90.20 & \textcolor{gray}{25.97}/85.34 & \textcolor{gray}{25.88}/83.02 & \textcolor{gray}{25.60}/81.53 & \textcolor{gray}{25.55}/\textbf{81.16} \\
 & $\mathcal{L}_c$ & \textcolor{gray}{2.07}/2.22 & \textcolor{gray}{1.48}/2.19 & \textcolor{gray}{0.30}/2.15 & \textcolor{gray}{0.04}/2.12 & \textcolor{gray}{0.01}/2.11 & \textcolor{gray}{0.01}/2.12 & \textcolor{gray}{0.01}/2.14 \\
 & $\mathcal{L}_r$ & \textcolor{gray}{0.84}/0.24 & \textcolor{gray}{2.35}/0.10 & \textcolor{gray}{6.46}/0.06 & \textcolor{gray}{9.25}/0.03 & \textcolor{gray}{10.40}/0.01 & \textcolor{gray}{11.06}/0.01 & \textcolor{gray}{11.51}/0.00 \\
\cline{1-9} \multirow{3}{*}{DO~\cite{srivastava2014dropout}} & TA & \textcolor{gray}{96.13}/94.43 & \textcolor{gray}{96.47}/95.03 & \textcolor{gray}{95.93}/95.03 & \textcolor{gray}{92.74}/94.79 & \textcolor{gray}{81.96}/92.15 & \textcolor{gray}{68.12}/92.69 & \textcolor{gray}{55.39}/\textbf{91.70} \\
 & $\mathcal{L}_c$ & \textcolor{gray}{2.22}/2.23 & \textcolor{gray}{2.20}/2.22 & \textcolor{gray}{2.17}/2.20 & \textcolor{gray}{2.13}/2.20 & \textcolor{gray}{2.05}/2.20 & \textcolor{gray}{1.94}/2.21 & \textcolor{gray}{1.79}/2.23 \\
 & $\mathcal{L}_r$ & \textcolor{gray}{0.26}/0.24 & \textcolor{gray}{0.30}/0.09 & \textcolor{gray}{0.41}/0.04 & \textcolor{gray}{0.61}/0.02 & \textcolor{gray}{1.00}/0.01 & \textcolor{gray}{1.50}/0.00 & \textcolor{gray}{1.92}/0.00 \\
\cline{1-9} \multirow{3}{*}{WD~\cite{krogh1992simple}} & TA & \textcolor{gray}{88.91}/95.21 & \textcolor{gray}{50.87}/95.47 & \textcolor{gray}{27.98}/95.17 & \textcolor{gray}{27.66}/94.03 & \textcolor{gray}{25.14}/91.42 & \textcolor{gray}{28.05}/89.81 & \textcolor{gray}{25.57}/\textbf{86.98} \\
 & $\mathcal{L}_c$ & \textcolor{gray}{2.16}/2.22 & \textcolor{gray}{1.87}/2.20 & \textcolor{gray}{1.06}/2.18 & \textcolor{gray}{0.32}/2.16 & \textcolor{gray}{0.07}/2.16 & \textcolor{gray}{0.04}/2.17 & \textcolor{gray}{0.02}/2.19 \\
 & $\mathcal{L}_r$ & \textcolor{gray}{2.94}/0.23 & \textcolor{gray}{10.52}/0.09 & \textcolor{gray}{26.04}/0.05 & \textcolor{gray}{53.65}/0.02 & \textcolor{gray}{81.53}/0.01 & \textcolor{gray}{84.64}/0.00 & \textcolor{gray}{107.80}/0.00 \\
\bottomrule
\end{tabular}
\caption{Test accuracy (TA), training cross-entropy loss $\mathcal{L}_c$, and our training regularization loss $\mathcal{L}_r$ which is shown for AL2 multiplied by 100 for readability. We evaluate all metrics at different epochs and with different baselines (no regularization Bare, batch normalization BN~\cite{ioffe2015batch}, dropout DO~\cite{srivastava2014dropout}, and weight decay WD~\cite{krogh1992simple}), without/with AL2. The networks are trained on the MNIST dataset with $75\%$ corrupt labels.}
\label{table:MNIST}
\end{table*}

\begin{table*}[t]
\centering
\begin{tabular}{c| *{6}{c}||c||}
\toprule
\multicolumn{8}{c}{\textbf{Area under cumulative ablation curve (/100) evaluated across training epochs (\textcolor{gray}{without}/with AL2)}} \\ \cline{1-8}
Baseline & epoch=100 & epoch=200 & epoch=300 & epoch=400 & epoch=500 & epoch=600 & epoch=700\\  \cline{1-8} 
{Bare} & \textcolor{gray}{35.44}/77.81 & \textcolor{gray}{19.17}/72.67 & \textcolor{gray}{15.52}/69.44 & \textcolor{gray}{14.73}/64.11 & \textcolor{gray}{15.36}/55.08 & \textcolor{gray}{15.36}/51.73 & \textcolor{gray}{15.19}/\textbf{47.65} \\
{BN~\cite{ioffe2015batch}} & \textcolor{gray}{35.08}/77.01 & \textcolor{gray}{19.17}/71.23 & \textcolor{gray}{15.80}/63.48 & \textcolor{gray}{15.79}/57.42 & \textcolor{gray}{15.64}/55.67 & \textcolor{gray}{15.69}/54.97 & \textcolor{gray}{15.60}/\textbf{54.96} \\
{DO~\cite{srivastava2014dropout}} & \textcolor{gray}{81.66}/78.52 & \textcolor{gray}{79.90}/78.74 & \textcolor{gray}{76.23}/78.80 & \textcolor{gray}{70.38}/78.31 & \textcolor{gray}{60.17}/73.57 & \textcolor{gray}{49.86}/73.30 & \textcolor{gray}{41.39}/\textbf{71.61} \\
{WD~\cite{krogh1992simple}} & \textcolor{gray}{39.50}/78.18 & \textcolor{gray}{20.74}/74.83 & \textcolor{gray}{15.94}/74.39 & \textcolor{gray}{16.09}/72.97 & \textcolor{gray}{15.40}/67.62 & \textcolor{gray}{16.12}/64.85 & \textcolor{gray}{15.35}/\textbf{62.63} \\
\bottomrule
\end{tabular}
\caption{We evaluate the area under the cumulative ablation curve at different epochs and with different baselines, without/with AL2. The networks are trained on the MNIST dataset with $75\%$ corrupt labels.}
\label{table:MNIST_75_AUC}
\end{table*}

\section{Label randomization experiments} \label{sec:evaluation}
We evaluate our method on a VGG-like (2D convolution, maxpooling, ReLU, and linear layers) CNN architecture\footnote{\href{https://github.com/majedelhelou/AL2}{https://github.com/majedelhelou/AL2}}, designed to examine the effects of different regularizers. For reproducibility purposes, all the details of the network architecture and the training settings are presented in the supplementary material\footnote{\href{https://infoscience.epfl.ch/record/271444}{https://infoscience.epfl.ch/record/271444}}. For weight decay, which is sensitive to its chosen weight, we run a parameter search and find the best weight decay value of $5\times 1e-4$, which is also a value typically used in practice. This value gives the best performance for weight decay without AL2.

We carry out the evaluation of network memorization with label randomization experiments~\cite{zhang2017understanding,neyshabur2017exploring}.
For each training dataset, a fixed percentage of labels is corrupted with labels chosen uniformly at random from the set of incorrect labels for a given training image (symmetric label noise). We then train the baseline (bare) network with no regularization, the network with batch normalization (BN), with dropout (DO) or with weight decay (WD) on the same corrupt dataset and starting from the same weight initialization. We repeat the training of each of these four networks with our AL2 regularization, again with the same corrupt dataset and starting from the same weight initialization. Results are reported in Table~\ref{table:MNIST} for the MNIST dataset and for $75\%$ corrupt random labels. Further results on MNIST, Fashion-MNIST, and CIFAR10 each with $75\%$, $50\%$, $25\%$, and $0\%$ corrupt random labels are additionally provided in the supplementary material, totaling 96 different networks trained for 700 epochs each.

The results in Table~\ref{table:MNIST} show the test accuracy (TA), the cross-entropy loss $\mathcal{L}_c$ and our regularization loss $\mathcal{L}_r$ at different epochs during the network training. We see that using AL2 significantly improves the generalization of the network assessed at the final epoch, by limiting the overfitting through regularization. The test accuracy improvement is of 60 percentage points in the most extreme case (with weight decay, Table~\ref{table:MNIST}). Without using AL2 during training, the best performance is obtained when using dropout. With dropout, compared to other network configurations, we note one interesting phenomenon. The network trained with dropout regularization also indirectly minimizes $\mathcal{L}_r$, an order of magnitude smaller than with batch normalization, and two order of magnitudes smaller than with weight decay or the bare network. We thus notice that dropout tends to lead to a smaller $\mathcal{L}_r$, which AL2 explicitly penalizes to a much larger degree. Counter-intuitively, weight decay hardly decreases the magnitude of the activations in the final feature representation layer that is created by the trunk of the network. This underlines the different effects obtained by regularizing activations or regularizing network weights as done by weight decay. 

These observations and insights on experiments with no corrupt labels are discussed in more detail in the supplementary material. We also note here that on $75\%$ corrupt data, the bare baseline achieves a test accuracy of $25.84\%$ but of $68.46\%$ with AL2, while the training cross-entropy loss is non-zero for the AL2-regularized network. This indicates that we could correct the labels by re-labeling the dataset with the AL2-regularized network then repeat the training to improve the performance. However, our objective is not to classify data with noisy-label training, but rather to use the randomized label tests to assess network generalization against memorization strength, for different regularizers.

\section{Neural Representation Analysis}

\begin{figure*}[t]
	\centering
	\includegraphics[width=\linewidth]{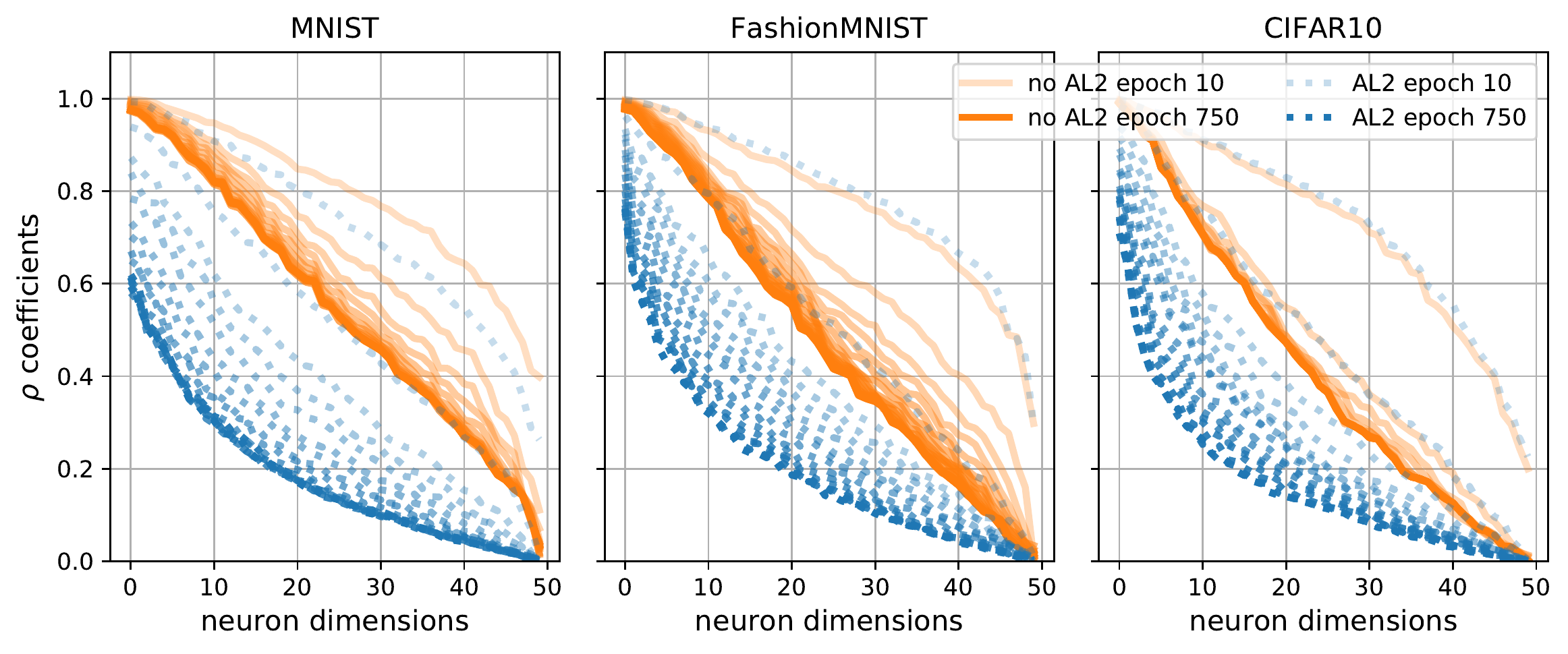}
	\caption{Canonical correlation coefficients $\rho$ as a function of neuron dimensions between the learned feature representation $\phi(x)$ at the beginning of training and at given training epochs ranging from 10 to 750 (illustrated with increasing color intensities), for the three datasets MNIST, FashionMNIST and CIFAR10. The plots show that AL2 has a significant effect on the representation learning process, confirming the fundamentally different classification results reported in Table~\ref{table:MNIST}. Results are obtained with a training dropout rate of 50 percent and for the first batch of each dataset. Both networks are initialized with identical weight values for a fair comparison. Best viewed on screen.}
    \label{fig:cca}
\end{figure*}

\subsection{Representation analysis with canonical correlation} \label{subsec:CCA}

The representation learned by a neural network depicts how different neurons respond to the given input data, in terms of their activation values. We feed forward a data point to the network and collect the activations of all neurons in a given layer into a vector in $\mathbb{R}^a$. Collecting and grouping such vectors for $n$ different data points yields the matrix $R_1 \in \mathbb{R}^{a \times n}$, which holds the representation of that set of data points by the neural network. Recent methods have proposed to use canonical correlation by computing a similarity metric from the series of correlation coefficients, which we briefly review in what follows.

The canonical correlation coefficient $\rho$ for two matrices $R_1 \in \mathbb{R}^{a \times n}$ and $R_2  \in \mathbb{R}^{b \times n}$ is given by
\begin{equation}
    \rho = \underset{(\omega_1, \omega_2) \in (\mathbb{R}^{a}, \mathbb{R}^{b}) }{max} \Bigg( \frac{ \langle \omega_1^T R_1, \omega_2^T R_2 \rangle }{ || \omega_1^T R_1 || \cdot || \omega_2^T R_2 ||} \Bigg),
\end{equation}
and the corresponding canonical correlation directions are $\omega_1^T R_1$ and $\omega_2^T R_2$. $\rho$ measures the degree of correlation between these two direction vectors. One can solve for the next-best $\rho$ value, which corresponds to two new direction vectors $\omega_1^T R_1$ and $\omega_2^T R_2$ that are respectively orthogonal to the corresponding first two vectors. Repeating this process, with each vector of the new couple ($\omega_1^T R_1$, $\omega_2^T R_2$) being orthogonal to the vector space spanned by the corresponding previously-found direction vectors, yields a sequence of $\rho$ values of size $min(a,b)$. These coefficient values are indicative of the similarity between $R_1$ and $R_2$. The larger the values are, the more similar are $R_1$ and $R_2$.

Both SVCCA~\cite{raghu2017svcca} and PWCCA~\cite{morcos2018insights} compute weighted averages of the canonical correlation coefficients. To avoid any loss of information through averaging, we visualize the entire sequences of $\rho$ coefficients in our analysis (Fig.~\ref{fig:cca}). 

For each dataset, we obtain a representative sample of correlation coefficients by passing the first random training batch through the network. We form a matrix of shape 50 $\times$ 16,000 for MNIST and FashionMNIST and 50 $\times$ 25,000 for CIFAR10. These matrices consist of the flattened activations at the layer before classification, i.e. the intermediate activations $\phi(x)$. We can thus obtain 50 correlation coefficients for the 50 neuron dimensions. We repeat this process at different training epochs, and compare the representation at a given epoch with the initial one. Note that all compared networks, without and with AL2, are initialized with the same set of random weights for a fair comparison. Since CCA is scale-invariant, this metric only depicts structural similarities between representations and can thus provide a good insight into the representation progress. A simple scaling down of the activation values does not affect the similarity measure. Figure \ref{fig:cca} therefore shows that AL2 significantly modifies the learning and the final learned representations. As supported by the results reported in Section~\ref{sec:evaluation}, including our regularization thus pushes the network's learning towards a fundamentally different representation, reducing the effect of overfitting.

\subsection{Generalization analysis with cumulative ablations} \label{subsec:ablation}
Analyzing generalization can also be carried out with a different approach than randomization experiments. A recent approach shows the correlation between network generalization and the area under the cumulative ablation curve~\cite{morcos2018importance}. This cumulative ablation curve is defined by the authors as the accuracy of the pre-trained network for different percentages of ablations going from zero to $100\%$, on the training set. An ablation of $20\%$ consists of systematically setting $20\%$ of the activations in the feature representation layer to zero during the feed-forward inference of the pre-trained network. These ablations are also said by the authors to be related to sharpness~\cite{keskar2017large}, which is found to be a good indicator of generalization strength when combined with a norm metric~\cite{neyshabur2017exploring}. 

We apply cumulative ablations on our pre-trained networks and report results in Table~\ref{table:MNIST_75_AUC}. At inference time, the activations of the feature representation layer obtained with the trunk $\phi(\cdot)$ of the network are set to zero at increasing rates going from zero to $100\%$ in steps of $10$. We measure the accuracy with each of the ablation rates, and calculate the area under the curve. We repeat this procedure at an interval of 100 epochs for each of the 8 networks to create the results of Table~\ref{table:MNIST_75_AUC}. The generalization assessment results are consistent with those discussed in Section~\ref{sec:evaluation}, and confirm our previous observations (this is the case across our diverse experiments and datasets). We also note that even between epochs 300 and 400, where the dropout network still does not overfit and performs similarly without and with AL2 (Table~\ref{table:MNIST}), the area under the cumulative ablation curve is larger with AL2 training as shown in Table~\ref{table:MNIST_75_AUC}.


\section{Conclusion} \label{sec:conclusion}
We propose a novel progressive activation loss (AL2) to regularize neural networks. Our loss acts increasingly with epochs on the magnitude of the activation signals of the feature representation layer. We use canonical correlation analysis to study the effect of AL2 on the learned feature representation throughout the training. This shows empirically the significant effect of our regularization on the fundamental representation that is learned by the networks.

We analyze memorization and generalization with randomization tests and with a cumulative ablation study to show the improvements of our AL2 method over state-of-the-art regularization techniques on three standard benchmark datasets.


\vfill\pagebreak

\small
\bibliographystyle{IEEEbib}
\bibliography{refs_shrink}

\end{document}